%% file: root.tex
\pdfoutput=1

\documentclass[journal]{IEEEtran}
\usepackage{graphics}
\usepackage{graphicx}
\usepackage{amsmath} 
\usepackage{amssymb}  
\usepackage{algorithm}
\usepackage{algpseudocode}
\usepackage{algorithmicx}
\usepackage{subfigure}
\usepackage{url}
\usepackage{color}
\ifCLASSINFOpdf
\else
\fi

\newcommand\kevinupdate[1]{\textcolor{black}{#1}}

\begin{document}
%
\title{Robot Vision Architecture for Autonomous Clothes Manipulation}
%
%
%

\author{Li Sun, Gerardo~Aragon-Camarasa, Simon~Rogers, J. Paul Siebert
\thanks{Li Sun was with the School of Computing Science, University of Glasgow, Glasgow, G12 8RZ, UK,
 e-mail: lisunsir@gmail.com}
\thanks{Manuscript received April 19, 2005; revised August 26, 2015.}}

%
%

\markboth{Journal of \LaTeX\ Class Files,~Vol.~14, No.~8, August~2015}%
{Shell \MakeLowercase{\textit{et al.}}: Bare Demo of IEEEtran.cls for IEEE Journals}
%



\maketitle

\begin{abstract}

This paper presents a novel robot vision architecture for perceiving generic 3D clothes configurations. Our architecture is hierarchically structured, starting from low-level curvatures, across mid-level geometric shapes \& topology descriptions; and finally approaching high-level semantic surface structure descriptions. We demonstrate our robot vision architecture in a customised dual-arm industrial robot with our self-designed, off-the-self stereo vision system, carrying out autonomous grasping and dual-arm flattening. It is worth noting that the proposed dual-arm flattening approach is unique among the state-of-the-art robot autonomous system, which is the major contribution of this paper. The experimental results show that the proposed dual-arm flattening using stereo vision system remarkably outperforms the single-arm flattening and widely-cited Kinect-based sensing system for dexterous manipulation tasks. In addition, the proposed grasping approach achieves satisfactory performance on grasping various kind of garments, verifying the capability of proposed visual perception architecture to be adapted to more than one clothing manipulation tasks.


\end{abstract}


%
\IEEEpeerreviewmaketitle

\section{Introduction}
\input{./introduction}

\section{Literature Review}\label{sec:related_work}
\input{./related_work}

\section{An Overall Schema of the Autonomous System}\label{sec:overall_schema}
\input{./overview}

\section{Hierarchical Vision Architecture}\label{sec:architecture}
\input{./visual_architecture}

\section{Garment Manipulation using the Proposed Visual Architecture}\label{sec:garment_manipulation}
\input{./overall}
\subsection{Heuristic Garment Grasping}\label{sec:grasping}
\input{./grasping}
\subsection{Dual-Arm Garment Flattening}\label{sec:flattening}
\input{./flattening}

\section{Experiments}\label{sec:chapter4_exp}
In this section, both the proposed grasping approach and flattening approach are evaluated in our integrated autonomous system. The grasping experimental result is presented in Section \ref{sec:chapter4_exp_1} and the dual-arm flattening is evaluated in Section \ref{sec:chapter4_exp_2}.
\input{./exp_overview}
\subsection{Garment Grasping Experiments}\label{sec:chapter4_exp_1}
\input{./grasping_exp}

\subsection{Garment Flattening Experiments}\label{sec:chapter4_exp_2}
\input{./flattening_exp}

\section{Conclusion}\label{sec:chapter4_conclusion}
\input{./conclusions}






%
\bibliographystyle{IEEEtran}
\bibliography{refs}

%




\end{document}

%% file: introduction.tex
Dexterous manipulation of clothing is a difficult task for autonomous robotic systems. This is because a robot needs to perceive and understand the state of a garment configuration. Current research efforts have been advocated to solve subtasks within an autonomous laundering pipeline, these are: grasping clothes from a heap of garments \cite{ramisa2012using,arnau2014finddd}, recognising clothes categories~\cite{ramisa2012using,willimon2011model,willimon2013classification}, unfolding \cite{2011bringing,willimon2011model,andreas2014autonomous,andreas2014active,li2015icra,triantafyllou2016geometric}, garment pose estimation \cite{kita2009clothes,kita2009method,li2014icra,li2014iros} and folding \cite{maitin2010cloth,van2011gravity,miller2012geometric,jan2014garment,li2015iros}. 

As further discussed in Section \ref{sec:related_work}, we found that existing approaches for cloth perception have been devised as ad-hoc robot vision solutions rather than generlisable robot vision architectures that can be adapted into different robotic clothes manipulation tasks. Likewise, garment flattening is a significant process within the laundering pipeline which is under-developed in the literature thus far (with exception to our early work in \cite{sun2013heuristic} and \cite{sun2015icra}). In previous research\cite{2011bringing}, the table and gravity is utilized for flattening. That is, conducting a sliding-table-edge flattening strategy in order to flatten a wrinkled garment. However, garment's wrinkles cannot be completely removed. Hence, to overcome the state-of-the-art limitations, we describe in this paper a novel robot vision architecture capable of perceiving and understanding deformable objects for successful and dexterous manipulation. Our architecture transforms low-level 3D visual features into rich semantic descriptions to underpin dexterous manipulation. We also present a binocular robot head able to provide accurate depth sensing in order to carry out autonomous dexterous grasping and dual-arm flattening manipulation robotic tasks.

\begin{figure}[t]
\centering
\includegraphics[width= 0.5\textwidth, natwidth=691, natheight=494]{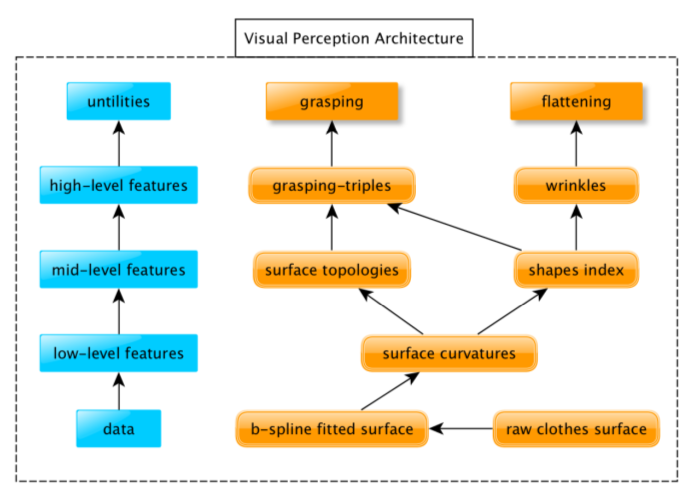}
\caption{The hierarchical visual architecture for visually-guided clothes manipulation. }
\label{fig:feature_architecture}
\end{figure}

More specifically, in order to offset these limitations of the predominant reported approaches, the main contributions of this paper are triple-fold: 

\begin{itemize}
\item \kevinupdate{Firstly, we firstly adapt an off-the-shelf, high-quality active binocular robot head to robot clothes perception and manipulation instead of using Kinect-like cameras. Our binocular robot head outperforms the depth sensing in both resolution and quality. In this paper, the binocular robot head incorporating a GPU stereo matcher tuned specifically for clothing provides accurate depth sensing for parsing the garment's 3D configuration. }

\item \kevinupdate{Secondly, a generic vision architecture is proposed which is capable of conducting multiple garment manipulation tasks. This proposed approach solves visually-guided garment manipulation tasks on the basis of sufficient parsing of the 3D configuration of the garment by the means of 3D surface shape and topology analysis. Thus, a bottom-to-top visual architecture is proposed to achieve a full understanding of the 3D garment's configuration. Consequently, through parsing the garment's configuration, the precise grasping and flattening manipulating skills can be indicated from visual guidances.}

\item \kevinupdate{Thirdly, the proposed vision architecture is integrated with dual-arm robot and a novel autonomous dual-arm flattening system is achieved. To the best of our knowledge, we firstly demonstrated a visually-guided, fully-autonomous, dual-arm flattening system, which can flatten various type of wrinkled garments placed on a table. A video demo can be found at: \url{http://youtu.be/Z85bW6QqdMI}.}
\end{itemize}

The structure of this paper is: In Section \ref{sec:related_work}, a comprehensive literature review is given which reviews the state-of-the-art achievements of visually-guided garment manipulation. Section \ref{sec:overall_schema} provides an overall schema of our autonomous clothes manipulation system including our customised robot, stereo robot head, the hand-eye calibration and the proposed visual architecture. In Section \ref{sec:architecture}, the hierarchical visual architecture for generic garment surface analysis is detailed. Section \ref{sec:garment_manipulation} presents the proposed visually-guided grasping approach and dual-arm flattening approach, which are on the basis of proposed visual perception architecture. The experimental validations  of the proposed autonomous grasping and flattening are detailed in Section \ref{sec:chapter4_exp}. The conclusion of this work is given in Section \ref{sec:chapter4_conclusion}.



%% file: related_work.tex
This section provides an overview on previously-reported robot vision approaches that have been employed for dexterous clothes manipulation tasks. These tasks include clothes grasping, unfolding, folding and flattening. 

\subsection{Visually-guided Garment Manipulations}
\subsubsection{Garment Grasping}
Ramisa et al. \cite{ramisa2012using,arnau2014finddd} proposed a grasping position detection approach using RGBD data. In their approach, SIFT and GDH (Geodesic Distance Histogram) local features are extracted on wrinkled regions in the RGB and depth domain, respectively. After Bag-of-Features coding, two layers of SVM classifiers are trained with linear and $\chi^{2}$ kernels. During the testing phase, a sliding window method is employed to detect graspable positions. After detection, `wrinkledness' measurements calculated from the surface normals are used to select the best grasping position.

The main limitation of the above approaches is that SVM-like sparse classifiers are not able to quantify the quality of grasping predictions. Moreover, training classifiers over human-labelled examples is biased to the judgement of the person labelling; for example, dissimilaraties between humans' and robot's end effectors are not taken into account while providing samples. \kevinupdate{Hence, the robot's ability to fully characterise the shape types and topology of cloth's surfaces is necessary in order to automatically detect potential grasping candidates and evaluate the goodness of those candidates with the consideration of its own body and visual limitations}. 

\subsubsection{Garment Unfolding}
The key step for garment unfolding is to detect grasping locations that can pottentially lead to a unfolding state (e.g. corners of a towel, shoulders of a shirt, waist of a pant, etc.). In this case, Cusumano et al. \cite{2011bringing} have proposed a multi-view based detection approach for unfolding towels. Their technique is based on finding two corners that are along the same edge of a towel. Following on Cusamano's work, Willimon et al. \cite{willimon2011model} proposed an interactive perception-based strategy to unfold a towel on the table. Their approach relies on detecting depth discontinuities on corners of towels. For each iteration, the highest depth corner on the towel is grasped and pulled away from its centre of mass. This approach is constrained to a specific shape of cloth (square towel), hence it is unlikely to be extended to other clothing shapes. 

Doumanoglou et al. \cite{andreas2014active,andreas2014autonomous} have proposed a general unfolding approach for all categories of clothes. Their approach is based on active random forests and hough forests which are used to detect grasping positions on hanging garments. Unfolding is carried out by iteratively grasping the lowest point of the observed garment until an unfolding stage is detected. In the following work \cite{triantafyllou2016geometric}, geometry clue e.g. edges of depth map, is employed to detect the grasping positions for unfolding. Li et al. have also devised an interactive unfolding strategy \cite{li2015icra}, in which the importance of grasping positions for unfolding are modelled based on Gaussian density functions.

\subsubsection{Garment Folding}
Miller et al. \cite{miller2011parametrized} modelled each category of clothes with a parametric polygonal model. They proposed an optimisation approach to approximate a polygonal models based on the 2D contours of clothing obtained after segmenting garment from the backgroud. Thereafter, this approach was used to fold garments based on a gravity-based folding \cite{van2011gravity} and geometry-based folding \cite{miller2012geometric} strategies. In Stria et al.'s method \cite{jan2014garment}, the key points on the contour (i.e. collar points, sleeve points, .etc) and match with the polygonal model, thereby accelerating the matching procedure.

The parametric polygonal models are unlikely to be adapted to other laundering tasks. This is due to the fact that polygonal models cannot recognise or track the state of a garment during folding, i.e. the state of the clothes are discretised into several steps during folding and the transition between states are assumed to be known. Likewise, only 2D contours are considered, and therefore, visual knowledge is insufficient for other laundering tasks that need to perceive the garment 3D surface.

\subsubsection{Interactive Perception in Garment Manipulation}
Interactive perception has been playing a critical role in dexterous clothing manipulaton. That is, a robot iteratively progress from an unrecognisable or initial state towards its recognition or manipulation goal on which the perception-manipulation cycles are iteractively performed. More specifically, Willimon et al. \cite{willimon2011classification} first proposed to recognise the clothing's category from hanging configurations. In their approach, the hanging garment is interactively observed as it is rotated. In Cusumano et al. \cite{2011bringing}, the robot is drived to hang the garment and slid along the table edges (on both left and right side) iteratively until the robot can recognise its configuration and then to a unfolded configuration. Subsequently, in Doumanoglou et al.'s unfolding work \cite{andreas2014active}, an active forest is employed to rotate the hanging garment to a perceptually-confident field of view. Li et al. \cite{li2015icra} proposed a more straight-forward unfolding approach based on pose estimation \cite{li2014icra,li2014iros} through interactively moving the grasping point towards specific target positions (e.g. elbows). Moreover, interactive perception has been used in heuristic-based generic clothing manipulation.

In Willimon et al. \cite{willimon2011model} and our previous work \cite{sun2013heuristic,sun2015icra}, perception-manipulation cycles are conducted to track the flattening state of the garment and heuristic manipulation strategies are used to flatten the wrinkled garment on the operating table.

\subsubsection{Garment Flattening}
For the specific garment flattening robotic task tackled in this paper, current research falls into two categories: \textit{gravity-based flattening} \cite{andreas2014autonomous,andreas2014active} i.e. hanging the garment for reducing the wrinkledness or \textit{by sliding the garment along a table} \cite{2011bringing}. Since no visual-feedbacks are used in these methods, the garment cannot be guaranteed to be flattened.

\subsection{Limitations of the State-of-the-Art}

\begin{figure}[thpb]
  \centering
	\includegraphics[width= 0.49\textwidth, natwidth=697,natheight=399]{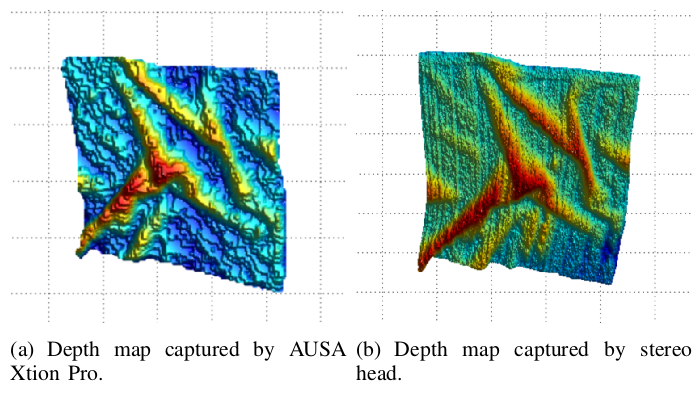}
    \caption{The comparison between depth data produced by kinect-like camera and stereo head.}
    \label{fig:rh_xtion}
\end{figure}

Form the literature review about visually-guided clothes manipulation, Kinect-like cameras are widely-used for cloth perception \cite{ramisa2012using,arnau2014finddd,andreas2014active,andreas2014autonomous,li2014icra,li2014iros,li2015icra,li2015iros}, and various manipulation approaches are proposed for resolving clothes gasping, unfolding and folding problems. Kinect-like cameras can provide real-time depth sensing with an precision of approximately 0.3-3cm\footnote{\kevinupdate{This depth sensing precision is depending on the range between camera and object. In our robot scenario, the precision is approximately 0.5-1.0cm. }}. Under such precision, the Kinect-like cameras can barely capture the small landmarks or estimate the magnitude of bending of clothes surfaces accurately, which are required for dexterous manipulations such as grasping and flattening. Moreover, reported methods are constrained to a specific garment or task at hand. In other words, the state-of-the-art visual perception approaches for clothes manipulation are not generic enough for more than one tasks. This can be attributed to the lack of sufficient understanding and perception of the clothes configuration, by which the generic landmarks can be localised and parametrised. To the best of our knowledge, Ramisa et al. \cite{ramisa2013finddd} proposed a 3D descriptor that is exploited in clothes grasping, wrinkle detection, and category recognition tasks, which is the only generic approach for multiple clothes perception tasks. However, all tasks rely heavily on classifiers which turned out to achieve limited performance. Likewise, they evaluate their manipulation performance in annotated datasets as opposed to real-life experiments where camera and robot calibration, mechanical and sensing errors and mistakes caused by labelling are not considered.

We can conclude that, the state-of-the-art visually-guided dexterous clothes manipulation have the following limitations: Firstly, Kinect-like and low-resolution depth cameras are not precise enough to sense garment details and as a consequence dexterous visually-guided manipulations become challenging. These types of cameras therefore constrain the application scope and capabilities of robots. Secondly, existing approaches for visually-guided clothes manipulation usually focus on specific tasks rather than parsing garments' geometrical configuration with sufficient understanding of surface shapes and topologies. As a consequence, most of the existing approaches are unlikely to be extended to multiple laundering tasks.

%% file: overview.tex
\begin{figure}[thpb]
\centering
\includegraphics[width= 0.49\textwidth,natwidth=671,natheight=497]{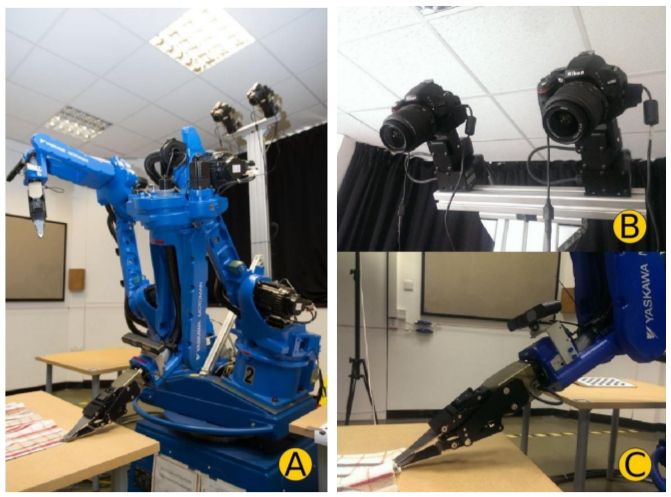}
\caption{(A) The CloPeMa robot which consists of two seven degrees of freedom YASKAWA arms and a custom made YASKAWA turn-table. Each arm features a specialised gripper for handling clothing and a ASUS Xtion Pro. (B) Our stereo robot head instagrated on the CloPeMa testbed suit. (C) A close up of the CloPeMa gripper.}
\label{fig:robot}
\end{figure}

\kevinupdate{In this section, a brief introduction of our proposed autonomous system is given. This system consists of the \textit{customized dual-arm robot}, \textit{stereo robot head}, \textit{stereo head calibration}, \textit{stereo matcher}, \textit{vision architecture}, \textit{visually-guided manipulation skills} and \textit{robot motion control}. The proposed vision architecture and visually-guided manipulation skills are the main contribution of this paper, which will be presented in Section \ref{sec:architecture} and Section \ref{sec:garment_manipulation}.}



\begin{figure*}[t]
\centering
\includegraphics[width= .9\textwidth,natwidth=1440,natheight=1660]{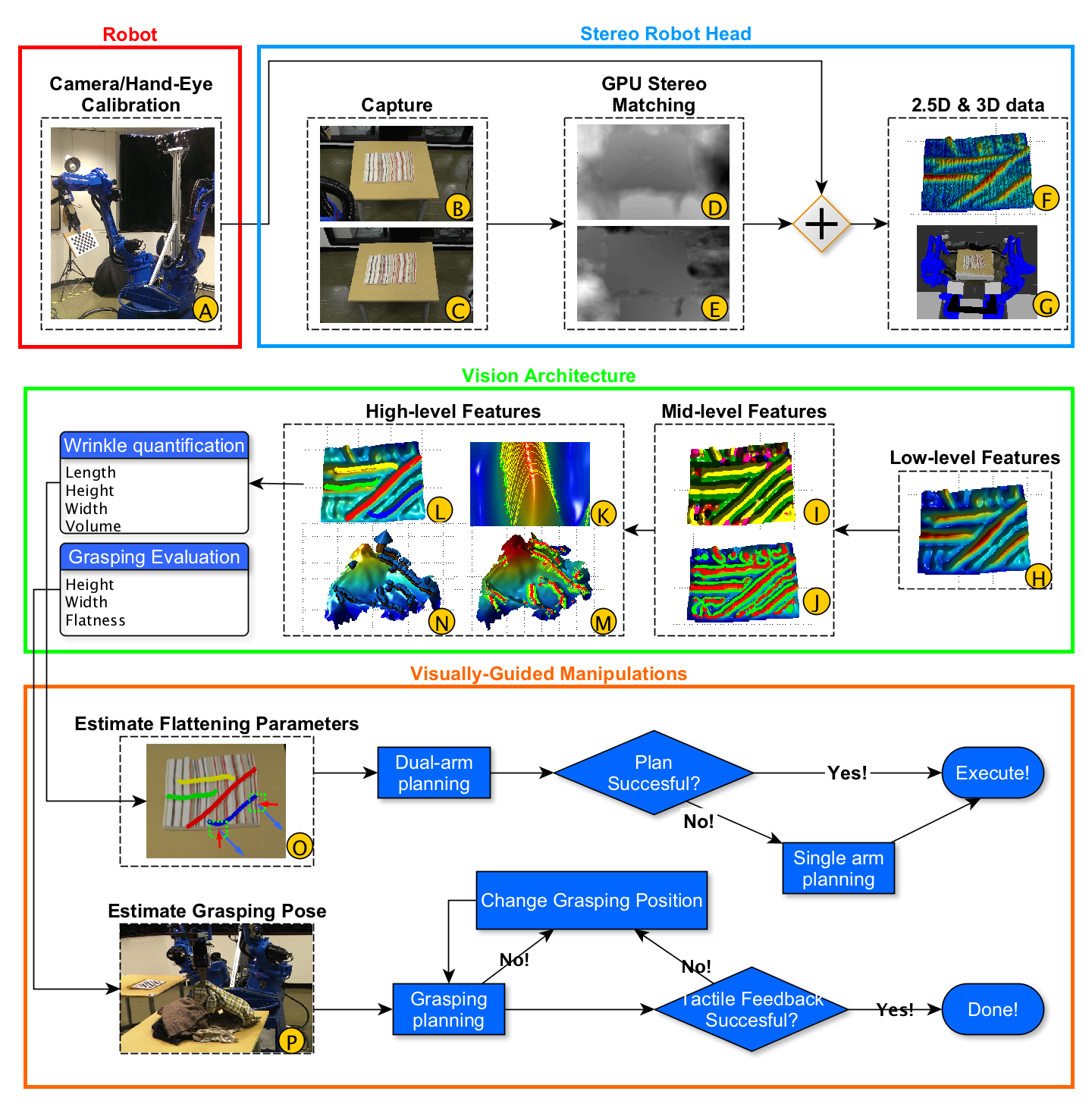}
\caption{\label{fig:pipeline_manipulation}The whole pipeline for autonomous grasping and flattening.}
\end{figure*}

\subsection{CloPeMa Robot}
The main robot manipulators are based on the industrial robotic components for welding operation which are supplied by YASKAWA Motoman. As shown in Fig. \ref{fig:robot}-A, two MA1400 manipulators are used as two robot arms. Each manipulator is of 6 DOF, 4 kg maximal load weight, 1434 mm maximal reaching distance, $\pm$0.08 mm accuracy. These specifications satisfy the requirements for conducting accurate adult clothing manipulation. They are mounted on rotatable turning tables. The robot arms and turning table are powered and controlled by DX100 controllers.

\subsection{CloPeMa Robot Head}
Differing from most of the state-of-the-art visually-guided manipulation research, we aims to use relatively inexpensive, commercially available component elements to build an off-the-shelf robot vision system (binocular head) for depth sensing. 

In order to offset the limitation of widely-used depth sensor such as Kinect w.r.t. accuracy and resolution, an off-the-shelf robot head is used in this paper for depth sensing. As shown in Fig. \ref{fig:robot}-B, the robot head comprises two Nikon DSLR cameras (D5100) that are able to capture images of 16 mega pixels \kevinupdate{through USB control. Gphoto library\footnote{http://gphoto.sourceforge.net/} is employed to drive the capturing under ubuntu.} These are mounted on two pan and tilt units (PTU-D46) with their corresponding controllers. The cameras are separated by a pre-defined baseline for optimal stereo capturing. Its field of view covers the robot work-space. The robot head provides the robot system with high resolution 3D point cloud.

\subsection{Stereo Head Calibration}
Our stereo head calibration has two steps: camera calibration and hand-eye calibration. The former is employed to estimate the intrinsic parameters of the stereo cameras. For the CloPeMa robot, the OpenCV calibration routines\footnote{\url{http//opencv.org}} are employed to estimate the intrinsic camera parameters of each camera. Furthermore, hand-eye calibration is employed to link the stereo head's reference frame into the robot kinematic chain. In other words, the unknown transformation from the camera frame to the calibration pattern coordinate system, as well as the transformation from the calibration pattern coordinate system to the hand coordinate system, need to be estimated simultaneously. For the CloPeMa stereo head, Tsai's hand-eye calibration \cite{tsai1988real,tsai1989new} routines are used to estimate rigid geometric transformations between camera to chess board and chess board to the gripper.

\subsection{Stereo Matcher}
Having calibrated and integrated the stereo-head, the next stage is stereo-matching and 3D reconstruction. In this procedure, a pair of images are captured simultaneously by the left and right cameras. The C3D matcher \cite{siebert1995c3d,zhengping1988multi} is employed to find the horizontal and vertical disparities of the two images. In the implementation for the CloPeMa robot head, C3D matcher is accelerated by CUDA\footnote{https://developer.nvidia.com/cuda-zone} GPU paralleling programming \cite{enlighten72079} to produce a 16 mega-pixel depth map in 0.2 fps. A GMM-based grab-cut \cite{stria14taros} pre-trained by table color information is employed to detect and segment the garment.

\subsection{Robot Motion Control}
The ClopeMa robot is fully integrated with Robot Operating System through ROS industrial package\footnote{ \url{http://wiki.ros.org/Industrial}}. More specifically, the \textit{URDF} (uniform robot description form) is used to define the geometric structure of the robot. After the geometric structure is defined, collision can be detected by robot collision models, and the transforms between robot links can be achieved by TF\footnote{ \url{http://wiki.ros.org/tf}}. \textit{MoveIt} package is employed to achieve the communication between user interface and robot controllers. 

%% file: visual_architecture.tex
This section presents the proposed vision architecture in details. Firstly, a piece-wise $B$-Spline surface fitting is adapted as pre-processing in Section~\ref{sec:featureframework_preprossing}, and the low-level feature extraction is presented in Section~\ref{sec:featureframework_low_feat}.  In Section~\ref{sec:_featureframework_mid_feat}, surface shapes and topologies are introduced as the mid-level features. Afterwards, two high-level features i.e grasping triplets and wrinkle description, are reported in Section~\ref{sec:featureframework_high_feat1} and Section~\ref{sec:featureframework_high_feat2}.

\subsection{Pre-Processing: B-Spline Surface Fitting}\label{sec:featureframework_preprossing}
As geometry-based 2.5D features such as curvatures and shape index are extremely sensitive to high frequency noise, a piece-wise B-Spline surface approximation is used to fit a continuous implicit surface onto the original depth map. More details are presented in our previous work \cite{sun2016ijars}.

\subsection{Low-Level Feature: Surface Curvatures Estimation}\label{sec:featureframework_low_feat}
To compute curvatures from depth, 2.5D points in the depth map (i.e. $x$, $y$ and depth $-$ $x$ and $y$ are in pixels while depth is in metres) are examined pixel by pixel in order to find if they are the positive extrema along the maximal curvature direction. That is, given a depth map $I$, for each point $p$ in $I$, the mean curvature $C_{m}^{p}$ and Gaussian curvature $C_{g}^{p}$ are firstly calculated by Eq. \ref{eq:m_curvature} and Eq. \ref{eq:g_curvature}, where first derivatives $f_{x}^{p}$, $f_{y}^{p}$, and second derivatives $f_{xx}^{p}$, $f_{yy}^{p}$,$f_{xy}^{p}$ are estimated by Gaussian convolution. Then, the maximal curvature $k_{max}$ and minimal curvature $k_{min}$ can be calculated by $C_{m}^{p}$ and $C_{g}^{p}$ (shown in Eq. \ref{eq:max_min_curvatures}). 

\begin{equation}
C_{m}^{p} = \frac{(1+(f_{y}^{p})^{2}) f_{xx}^{p} + (1+(f_{x}^{p})^{2}) f_{yy}^{p} - 2f_{x}^{p}f_{y}^{p}f_{xy}^{p}}{2(\sqrt{1+(f_{x}^{p})^{2} + (f_{y}^{p})^{2}})^{3}} 
\label{eq:m_curvature}
\end{equation}

\begin{equation}
C_{g}^{p} = \frac{f_{xx}^{p}f_{yy}^{p}-(f_{xy}^{p})^{2}}{(1+(f_{x}^{p})^{2}+(f_{y}^{p})^{2})^{2}} 
\label{eq:g_curvature}
\end{equation}

\begin{equation}
k_{max}^{p},k_{min}^{p} = C_{m}^{p} \pm  \sqrt{(C_{m}^{p})^{2}-C_{g}^{p}}\label{eq:max_min_curvatures}
\end{equation}

\subsection{Mid-Level Features: Surface Shapes and Topologies}\label{sec:_featureframework_mid_feat}


\subsubsection{Surface Shape Analysis} \label{sec:shape_analysis}

Shape index~\cite{koenderink1992surface}, performs a continuous classification of the local shape within a surface regions into real-value index values, in the range [-1,1]. Given a shape index map $S$, the shape index value $S^{p}$ of point $p$ can be calculated as follows \cite{koenderink1992surface}:
\begin{equation}
\centering
\displaystyle S^{p}= \frac{2}{\pi} \tan^{-1} \left [ \frac {k_{min}^{p}+k_{max}^{p}}{k_{min}^{p}-k_{max}^{p}} \right ],
\label{eq:shape_index}
\end{equation}
where $k_{min}^{p}$, $k_{max}^{p}$ are the minimal and maximal curvatures at point $p$ computed using Eq.\ref{eq:max_min_curvatures}. The shape index value is quantised into nine uniform intervals corresponding to nine surface types -- \textit{cup}, \textit{trough}, \textit{rut}, \textit{saddle rut}, \textit{saddle}, \textit{saddle ridge}, \textit{ridge}, \textit{dome} and \textit{cap}.

\begin{figure}[thpb]
\centering
\includegraphics[width= 0.49\textwidth,natwidth=700,natheight=367]{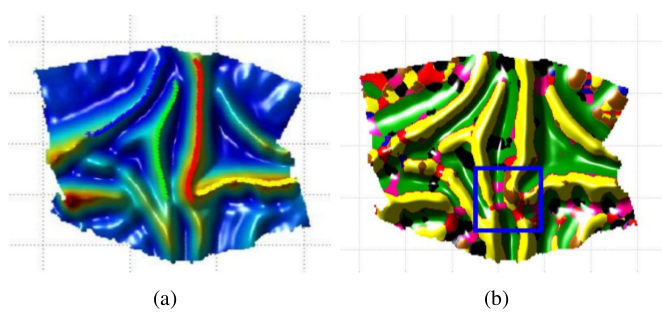}
\caption{An example of splitting wrinkle using Shape Index. In highly wrinkled situations, the shape of wrinkles at junctions are classified as dome or rut (as shown in brown and red colours); this classification is used to separate jointed wrinkles in this work.} 
\label{fig:highly_wrinkled}
\end{figure}

In order to parse the shape information exhibited by the visible cloth surface, the shape index map is calculated from each pixel of the depth map and a majority rank filtering is applied. This non-linear filtering removes outlier surface classifications and can be tuned to produce a relatively clean classification of shape types over the cloth surface. An example can be seen in Fig. \ref{fig:pipeline_manipulation}-I. It is worth noting that, the shape types `rut' and `dome' can be used to recognise the junction of multiple wrinkles thereby splitting wrinkles (as shown in Fig. \ref{fig:highly_wrinkled}).

\subsubsection{Surface Topologies Analysis}

Among all the shape types, \textit{ridges} is of critical importance in the analysis and description of wrinkles. In this paper, the definition of `ridges' shares similarities to that given by Ohtake et al. \cite{belyaev2005detection}. The main difference is that instead of estimating curvatures from a polygon mesh, surface curvatures are calculated using differential geometry, obtained directly from the depth map (as it is presented in Eq. \ref{eq:max_min_curvatures}).

The \textit{surface ridges} are therefore the positive extrema of maximal curvature while \textit{the wrinkle's contour} is the boundary of the concave and convex surfaces of the garment.  

From the nine shape types, four are convex (i.e. saddle ridge, ridge, dome, cap) and the rest are concave (i.e. cup, trough, rut, saddle rut, saddle). Thereby, the wrinkle's contour can be estimated. Alternatively, the boundary of the convex and concave surface can be more robustly estimated by computing the zero-crossing of the second derivatives of the garment's surface. In our implementation, a Laplace template window of size $16 \times 16$ is applied on the depth map in order to calculate the second order derivative. After the wrinkle's contour has been detected, the garment surface topologies are fully parsed. An example can be seen in Fig. \ref{fig:pipeline_manipulation}-J.

\subsection{High-Level Features - Grasping Triplets}\label{sec:featureframework_high_feat1}

In this paper, a wrinkle comprises a continuous ridge line localised within in a region where the surface shape type is `ridge'. The wrinkle is delimited (bounded) by two contour lines, each located on either side of the maximal curvature direction. A wrinkle can be quantified by means of a triplet comprising a ridge point and the two wrinkle contour points located on either side of the ridge, along the maximal curvature direction (as shown in Fig. \ref{fig:pipeline_manipulation}-K).

The above definition is inspired by classical geometric approaches for parsing 2.5D surface shapes and topologies (i.e. shape index \cite{koenderink1992surface}, surface ridges and wrinkle's contour lines). In this work, the height and width of a wrinkle are measured in terms of triplets. Accordingly, triplets can also be used as the atomic structures for finding and selecting grasping points (shown in Fig. \ref{fig:pipeline_manipulation}-N).

\subsubsection{Triplets Matching}\label{sec:triplet_matching}

From wrinkle's geometric definition, the maximal curvature direction $\theta$ can be calculated by Eq. \ref{eq:curvature_direction}. 
\begin{equation}
\centering
\theta = \tan^{-1}\frac{\partial y}{\partial x}.
\label{eq:curvature_direction}
\end{equation}
In this equation, $\partial y$ and $\partial x$ are the derivatives of $k_{max}$, computed by Gaussian convolution.

Given a ridge point $p_{r}$ in a depth map $I$ with scale $\varphi_{L1}$, this proposed method searches for the two corresponding contour points ($p_{c}^{l}$ and $p_{c}^{r}$) over the two directions defined by $\theta$ and its symmetric direction using a depth based gravity-decent strategy. If the searched path is traversed in the same `ridge' region as $p_{r}$ (shown as yellow in Fig. \ref{fig:pipeline_manipulation}-I), the process will continue. Otherwise, the searching will be terminated. Algorithmic details of triplet matching are described in our previous work \cite{sun2016ijars}.


Theoretically, every ridge point should be matched with its two corresponding wrinkle contour points. Whereas, due to occlusions and depth sensing errors, some wrinkle points fail to find their associated contour points and therefore do not generate a triplet. In order to eliminate the uncertainties caused by occlusions and errors, only triplets whose ridge points matched with both two wrinkle contour points are regarded as valid primitives for wrinkle quantification. An example of triplets matching is shown in Fig. \ref{fig:pipeline_manipulation}-K and M. Given a triplet $t_{p}$ containing one ridge point $p_{r}$ and two wrinkle contour points $p_{c}^{1}$ and $p_{c}^{2}$, the height $h_{t}$ and width $w_{t}$ can be calculated from the embedded triangle (triplets) using Eq. \ref{eq:triangle}. It is worth noting that, the triplet's points are transformed to the world coordinates, and as a consequence the unit of height $h_{t}$ and width $w_{t}$ is in meter.

\begin{equation}
\centering
\begin{array}{l}
 \displaystyle h_{t} = 2~ \frac{d \left( d - a \right) \left( d - b \right) \left( d - c \right)}{c}\\
 \displaystyle w_{t} = c,
 \end{array}
\label{eq:triangle}
\end{equation}
where $a$ = $\parallel p_{r},p_{c}^{1}\parallel_{2}$, $b$ = $\parallel p_{r},p_{c}^{2}\parallel_{2}$, $c$ = $\parallel p_{c}^{1},p_{c}^{2}\parallel_{2}$, and $d$ = $(a+b+c)/2$. The numerator of the right hand side of the equation is the area of a triangle embedded into the 3D space.


\subsection{High-Level Features: Wrinkle Description}\label{sec:featureframework_high_feat2}

\subsubsection{Wrinkle Detection\label{sec:wrinkle_det}}

The wrinkle detection consists of two steps: first, connecting ridge points into contiguous segments; second, grouping found segments into wrinkles (Fig. \ref{fig:pipeline_manipulation}-L). More details are given in our previous work \cite{sun2016ijars}.


After wrinkles have been detected, for each wrinkle, a fifth order polynomial curve is fitted along its ridge points. A high order polynomial curve is adopted in order to ensure that it has sufficient flexibility to meet the configuration of the wrinkles (here fifth order works well in practice). The polynomial curve denotes the parametric description of a wrinkle, and the curve function is defined as:

\begin{equation}
f(x)=ax^{5} + bx^{4} + cx^{3} + dx^{2} + ex + f,
\label{eq:wrikle_fun}
\end{equation}

\subsubsection{Hough Transform-Based Wrinkle Splitting}


\begin{figure}[thpb]
\centering
\includegraphics[width= 0.49\textwidth, natwidth=700,natheight=367]{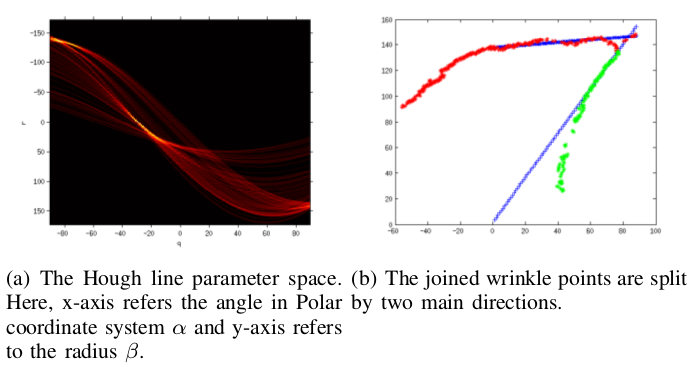}
\caption{\label{fig:hough_transform}\kevinupdate{Splitting joined wrinkles through Hough-Transform based wrinkle direction analysis. In Fig. (a), the two peak points refer to the two main directions of the joined wrinkles.  In Fig. (b), the two main hough line directions are plotted as blue line, and the points of joined wrinkle are split corresponding to these two direction, shown as red and green respectively.}}
\end{figure}

As reported in Section \ref{sec:shape_analysis}, \textit{Shape Index} is used to find junctions of wrinkles, where the shape types `rut' and `dome' are used as the visual cues for splitting wrinkles (as shown in Fig. \ref{fig:highly_wrinkled}). Furthermore, an additional Hough transform-based wrinkle splitting approach is proposed. \kevinupdate{In our approach, the joined wrinkles are parametrised as straight lines in Hough space in order to find the primary directions of the joined wrinkles.} Specifically, the Hough transform-based wrinkle splitting is used if the quality of wrinkle fitting (the \textit{RMSE} of the polynomial curve fitting above) is not acceptable. That is, each 2D point on the fitted wrinkle is projected as a curve in the Hough parameter space. Afterwards, peaks in the Hough space are ranked and the two largest peaks indicate the two main directions of the joined wrinkles. In order to avoid choosing two peaks originating from the same wrinkle, the two largest peaks should satisfy a non-locality constraint. In the implementation of this work, the value of 20 degrees works well in practice. Then, wrinkle points can be split into two subsets depending on the two largest peaks. \kevinupdate{An example of the proposed splitting is shown in Fig. \ref{fig:hough_transform}.} Finally, new polynomial curves are approximated on the split points respectively. This splitting procedure will be performed recursively until all the wrinkles are below an optimal \textit{RMSE} value (in practice, a value of 2 pixels works best for our implementation). Algorithm~\ref{alg:hough-transform} details the proposed Hough Transform based wrinkle splitting approach.

\begin{algorithm}[h]
\begin{algorithmic}[1]
\State \textbf{In:}
The detected wrinkles' points for splitting $\{P_{x}, P_{y}\}$, the threshold tolerance, $thres_{rmse}$, of the \textit{RMSE} wrinkle fitting, and non-locality constraints threshold, $thres_{\alpha}$.

\vspace{2mm}
\State \textbf{Out:}
The splited wrinkles' points $\{ {P_{x}^{1}, P_{y}^{1}} \}$ and $\{ {P_{x}^{2}, P_{y}^{2}} \}$.
\vspace{2mm}

\State
Approximate polynomial curve to $\{P_{x}, P_{y}\}$, and calculate the fitting error $rmse$.

\If {$rmse$ is larger than $thres_{rmse}$}
\State
Transform $\{P_{x}, P_{y}\}$ to Hough space, and get $\alpha$ and $\beta$ in Polar coordinate system
\State
Find the peak points in hough space and rank them w.r.t the accumulator values $\{ \hat{p}_{1}, ..., \hat{p}_{n_{p}} \}$.
\State
Find the two largest peak points $\hat{P}_{1}$ and $\hat{P}_{2}$ satisfying $\parallel \alpha_{\hat{P}_{1}}, \alpha_{\hat{P}_{2}}\parallel > thres_{\alpha}$.
\State
Restore two straight lines $l_{1}$ and $l_{2}$ in image space w.r.t two largest peaks in Hough space. 
\State 
Split the wrinkles' points $\{P_{x}, P_{y}\}$ into two subsets (${P_{x}^{1}, P_{y}^{1}}$ and ${P_{x}^{2}, P_{y}^{2}}$) through calculating the minimal \textit{Hausdorff} distances to $l_{1}$ and $l_{2}$.

\Else
\State
$\{P_{x}^{1}, P_{y}^{1}\}$ = $\{P_{x}, P_{y}\}$, and $\{P_{x}^{2}, P_{y}^{2}\}$ is empty.
\EndIf

\vspace{2mm}
\Return $\{P_{x}^{1}, P_{y}^{1}\}$ and $\{P_{x}^{2}, P_{y}^{2}\}$.
\end{algorithmic}
\caption{The Hough Transform based wrinkle splitting approach.}\label{alg:hough-transform}
\end{algorithm}

\subsubsection{Wrinkle Quantification}

\textit{Shape Index} classifies surface shapes without measuring surface magnitude. In our approach,  the magnitude of a wrinkle's surface is measured by means of triplets (as described in Section \ref{sec:triplet_matching}). Whereas, here the direction of triplet matching direction $\theta$ is estimated from the parametrised wrinkle description, which is more robust than that estimated from the maximal curvature direction. To be more specific, $\theta$ is computed from the perpendicular direction of the tangent line of the fitted curve on the observed wrinkle (i.e. fifth order polynomial curve in Eq. \ref{eq:wrikle_fun}). The tangent direction $\delta$ can be calculated as:

\begin{equation}
\delta = \arctan(5ax^{4}+ 4bx^{3} + 3cx^{2} + 2dx + e). 
 \label{eq:wrikle_tangent}
\end{equation}
 
Having obtained the searching direction $\theta$, the triplets on a detected wrinkle can be matched and thereafter their heights and width can be estimated by Eq. \ref{eq:triangle}. That is, given a wrinkle, $\omega$, containing a set of triplets $\{t_{1},...,t_{N_{r}}\}$, the width, $w_{w}$ and height, $h_{w}$ are calculated as the mean value of the width and height values of $\omega$'s triplets:

\begin{equation}
\centering
\begin{array}{l}
\displaystyle w_{\omega} = \sum_{t_{i}\in \omega}^{N_{r}} w_{t_{i}} / N_{t},$~~$
\displaystyle h_{\omega} = \sum_{t_{i}\in \omega}^{N_{r}} h_{t_{i}} / N_{t}\label{eq:wrinkle_quantification},
\end{array}
\end{equation}
where $\omega_{t_{i}}$ and $h_{t_{i}}$ are the width and height of the $i$th triplet of the wrinkle $\omega$.

For garment flattening, the physical volume of the wrinkle is adopted as the score for ranking detected wrinkles. \kevinupdate{PCA is applied on x-y coordinates of the largest wrinkle in order to infer the two grasping points and the flattening directions for each arm. More specifically, a 2 by 2 covariance matrix can be calculated from x and y coordinates, and then the principal direction of this wrinkle can be obtained by computing the eigenvector with respect to the largest eigenvalue.} To obtain the magnitude that the dual-arm robot should pull in order to remove the selected wrinkle, the geodesic distance between the two contour points of each triplet are estimated. Section~\ref{sec:flattening} details how these estimated parameters are used for flattening a garment.

%
%

%% file: overall.tex
This section presents the autonomous robot clothes manipulation systems with integrated visual perception architecture. The autonomous grasping approach is reported in Section \ref{sec:grasping} and dual-arm flattening is detailed in Section \ref{sec:flattening}.

%% file: grasping.tex
In this research, two visually-guided heuristic grasping strategies are proposed, in which the high-level grasping triplet feature (Section~\ref{sec:featureframework_high_feat1}) is adapted as the grasping location. Both strategies depend on an outlier removal strategy and grasping parametrisation for optimal garment manipulation as described in the following subsections.

\subsubsection{\kevinupdate{Central Wrinkle’s Points Estimation}}\label{sec:grasping_outline_removal}

\kevinupdate{Due to stereo matching errors caused by occlusions, inaccurate and incorrect topological descriptions may be detected, thereby affecting the estimations of grasping candidate. A central point evaluation mechanism is therefore devised to deal with isolated and inaccurate detections. This mechanism consists of computing the continuous hyper-exponential distribution of grasping triples.} That is, a \textit{Mahalanobis} distances based non-linear filtering is applied. Given a grasping triplet $t_{i}$ and the size of filter window\footnote{ From practical experience, a filter window of $32\times 32$ is used in our implementation.}, its \textit{Mahalanobis} distance can be calculated as follows:

\begin{equation}
\centering
D_{Mahalanobis}(p_{t_{i}},p_{T}) = \sqrt{(p_{t_{i}} - \mu_{T})^{T} \Sigma^{-1} (p_{t_{i}} - \mu_{T})}
\end{equation}

, where $p_{t_{i}}$ is the $x-y$ coordinate of $t_{i}$, $T$ refers to all the triples within the filter window, $\mu_{T}$ is the mean of the $x-y$ coordinates of all triples, and $\Sigma$ is the covariance matrix among all grasping triples within this region with respect to their spatial coordinates.

The probability of a grasping triplet being an outlier depends on the distance and direction within the spherical distribution. Hence, grasping triplets that are greater than a threshold\footnote{ From practical experience, a threshold of 0.5 is chosen in our implementation.} are treated as outliers and are removed from the list. This filtering is applied to every grasping triplet to probe whether it is an eligible grasping candidate.

\subsubsection{Grasping Parameter Estimation}\label{sec:grasping_parameters_estimation}

A \textit{good grasping position} is considered as where the grasping region is most likely to fit the gripper's shape (constrained by the robot joints limitations) and at the same time is most unlikely to change the garment's configuration when grasped. That is, the gripper must get grip of a large region of the clothing surface in order to provide a firm grasp on the clothes. In this approach, two robotic poses are required for a successful grasping action. These are: \textit{before-grasping} and \textit{after-grasping} poses. The \textit{before-grasping} pose is above the grasping point, while the \textit{after-grasping} pose indicates the lowest position the gripper should reach without colliding with the surface of the garment. By interpolating these poses sequentially, the robot is therefore able to conduct a smooth grasping action.


The required parameters for completing the two grasping poses mentioned above comprise: the before-grasping pose of the gripper with respect to the robot's world reference frame, the normal vector of grasping triplet and the rotation angle of the gripper with respect to the normal vector. The 3D position of gripper can be indicated by the detected grasping candidate. The grasping orientation of the gripper is set as the surface normal direction of the local region to grasp. In this paper, the surface normals are robustly estimated from the third principal direction of PCA of local point cloud. In order to obtain a robust estimation of the gripper rotation, the principal direction of graspable candidates within a local region is estimated and its perpendicular direction is used as the gripper rotation.

\subsubsection{Grasping Strategies}\label{sec:grasping_strategy}

In this paper, two grasping strategies are proposed: a \textit{height-priority} and a \textit{flatness-priority} strategy. For the height-priority strategy, the grasping energy of the motion of the gripper is minimised by selecting the candidates from the highest graspable points with respect to the robot's world reference frame. While the flatness-priority strategy computes a flatness score for each grasping candidate, $t$, that encodes the height, $h_{t}$, and the width, $w_{t}$ of the wrinkle's topology (Eq. \ref{eq:17}):

\begin{equation}
\centering
 \displaystyle flatness(t) =  \frac{h_{t}}{w_{t}} 
\label{eq:17}
\end{equation}

The height-priority strategy is able to grasp the clothing with the smallest cost of motion energy, and as a consequence the trajectory of planing is simpler, and is easier to solve the inverse kinematic problem and avoid collisions during motion planning. However, the drawback is also obvious, as the height-priority strategy cannot grantee that the mechanical shape of the gripper fits the region to grasp properly. In contrast, the flatness-priority strategy chooses the grasping candidate of the largest flatness rate, which is able to select the region most likely to fit the gripper but can bring difficulties to solving the inverse kinematic problem and avoiding collision. In our implementation, flatness-priority strategy and height-priority strategy are selected alternately until the grasping is completed.

%% file: flattening.tex
\subsubsection{Flattening Heuristic}


In this research, the heuristic flattening strategy adopts a greedy search approach, in which the largest wrinkle detected is eliminated in each perception-manipulation iteration. Only largest detected wrinkle is considered to be modified per iteration because the manipulation errors accumulated into the system increases when considering a group of wrinkles with similar directions, and the likelihood of applying appropriate flattening is significantly reduced. Therefore, the largest wrinkle detection heuristic guarantees that a solution is achieved regardless of highly wrinkled configurations. In our approach, wrinkles are quantified according to their physical volume in this chapter. The estimation of the volume of a wrinkle $w$ is given by:

\begin{equation}
\centering
volume_{w} = l_{r} * ((\sum_{t_{i}\in w}^{N_{r}} w_{t} \times h_{t}) / N_{t}),
\label{eq:16}
\end{equation}
where $N_{r}$ is the number of fitted ridge points in $w$, $N_{t}$ refers to the number of matched triplets, $t_{i}$ is the $i$th triplet of $w$, $w_{t}$ and $h_{t}$ refers to the width and height of the triplet $t_{i}$, and $l_{r}$ is the length of the wrinkle which is calculated by summing up the $L^{2}$ distances between every two nearest ridge points.

\subsubsection{Poses of a Primitive Flattening Action}\label{sec:a_flattening_action}

\begin{figure}[thpb]
\centering
    \includegraphics[width=0.35\textwidth,natwidth=497,natheight=347]{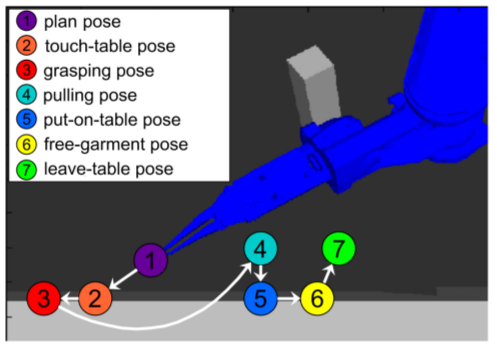}
    \caption{\label{fig:gripper_poses1} The seven poses for a robotic flattening motion. The gripper is moved to the `plan pose', from where the trajectory of gripper is interpolated among poses sequentially in order to move the gripper. It is noticeable that the grasping direction and pulling direction are not aligned. The \textit{plan pose}, \textit{touch-table pose} and \textit{grasping pose} are coplanar, while the \textit{grasping pose}, \textit{pulling pose}, \textit{put-on-table pose}, \textit{free-garment pose} and \textit{leave-table pose} are coplanar. For the gripper state, it will be set to `open' in \textit{plan pose}, `close' after \textit{grasping pose} and `open' again after \textit{put-on-table pose}. }
\end{figure}

\begin{figure}[thpb]
\centering
    \includegraphics[width=0.35\textwidth,natwidth=497,natheight=347]{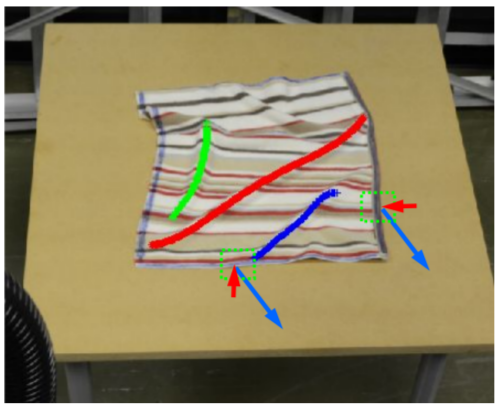}
    \caption{\label{fig:gripper_poses2} An example of detected wrinkles and the corresponding grasping poses and flattening directions of the dual-arms. The three largest wrinkles are shown, where the red one is the largest. The inferred grasping and flattening (pulling) directions are shown as red and blue arrows, respectively.}
\end{figure}
 
An entire flattening action consists of seven robotic poses: \textit{plan}, \textit{touch-table}, \textit{grasping}, \textit{pulling}, \textit{put-on-table}, \textit{free-garment} and \textit{leave-table}. These poses are illustrated in Fig. \ref{fig:gripper_poses1}. This figure also includes other pre-defined parameters used during the flattening task, e.g. orientation of the gripper w.r.t the table. The \textit{plan pose} (Fig. \ref{fig:gripper_poses1}, purple) refers to moving the robot's gripper close to the table by error-tolerance planing in preparation for flattening, then the gripper will approach the rest poses consequently by interpolating in \textit{Cartesian} coordinates system. The \textit{touch-table} and \textit{grasping poses} (Fig. \ref{fig:gripper_poses1}, orange and red, respectively) involve grasping the garment's boundary by interpolating the robot's motion between these two poses. The \textit{pulling} and \textit{put-on-table poses} pull the grasped garment according to the \textit{Geodesic} distance (Eq. \ref{eq:flattening_dist}) and smoothly return the garment to the table. Finally, the \textit{free-garment} and \textit{leave-table poses} are for freeing the garment and leaving the table.

In order to calculate and interpolate these robotic poses, four parameters are required: \textit{grasping position}, \textit{grasping direction}, \textit{flattening direction} and \textit{flattening distance}. The \textit{grasping} and \textit{pulling poses} are estimated using these parameters. Then the other poses are inferred from the \textit{grasping} and \textit{pulling poses}. By interpolating these seven poses sequentially, the robot is therefore able to perform a smooth flattening action. It worth noting that for planning and interpolation, the \textit{MoveIt} package is used\footnote{ Available in ROS: \url{http://moveit.ros.org}}.

\subsubsection{Flattening Parameters Estimation}
Here the details about how to set the four flattening parameters (described in Section \ref{sec:a_flattening_action}) are provided. As shown in Fig. \ref{fig:gripper_poses2}, once the largest wrinkle is selected, \textit{PCA} is employed to compute its primary direction, and the two \textit{flattening directions} are perpendicular to the wrinkle's primary direction. After the \textit{flattening directions} are fixed, the two corresponding cross points on the garment contour are set as the position of the \textit{grasping pose} (Fig. \ref{fig:gripper_poses1}). \kevinupdate{In single arm flattening, the intersection point refers to that between wrinkle’s bisector and garment's contour. Whereas, in dual-arm flattening, wrinkles are divided into two equal segments and the two intersection points are calculated respectively.} The \textit{grasping direction} is estimated by the local contours of the \textit{grasping positions} (as shown in Fig. \ref{fig:gripper_poses2}). The pulling distance $d_{w_{i}}$ of wrinkle $w_{i}$ is estimated by:
\begin{equation}
\centering
d_{w_{i}} = \sum_{t_{i}\in w}^{N_{r}} ( \emph{G}(c_{l}^{t_{i}},c_{r}^{t_{i}}) - \emph{E}(c_{l}^{t_{i}},c_{r}^{t_{i}}) ) / N_{t}) * Coeff_{spring},
\label{eq:flattening_dist}
\end{equation}
where $t_{i}$ is the $i$th $N_{r}$ triplet in $w_{i}$; $c_{l}^{t_{i}}$ and $c_{r}^{t_{i}}$ are its two wrinkle contour points; $\emph{G}$ refers to \textit{Geodesic} distance \cite{sethian1999level}\footnote{Gabriel Peyre's toolbox is used in the implementation of this work for calculating \textit{Geodesic} distance: \url{http://www.mathworks.co.uk/matlabcentral/fileexchange/6110-toolbox-fast-marching}}, while \emph{E} refers to \textit{Euclidean} distance. $Coeff_{spring}$ is the maximal distance constraint between particles in a mass-spring cloth model\footnote{ From practical experience, the $Coeff_{spring}$ is set as 1.10 in the experiments of this work.}.

\subsubsection{The Dual-Arm Collaboration}
Because of the limitation of the robot's joints and possible collisions between the two arms, not all of the \textit{planing poses} (the first pose of a flattening action) can be planned successfully. Therefore a greedy \textit{pose/motion exploration strategy} is proposed (The pseudo code of the proposed algorithm is provided in Algorithm \ref{alg:dual_arm_exploration}). This results in a significant improvement while flattening with both arms. However, if this algorithm fails, the robot only employs one arm; the arm used is selected according to the flattening direction\footnote{In order to enhance the success rate of motion planing, if the flattening action is towards left, then left arm is employed; otherwise, right arm is employed.}.

\begin{algorithm}[thpb]
\label{alg:dual_arm_exploration}
\begin{algorithmic}
\State \textbf{In:}
The direction interval is $d_{I}$. The maximum numbers of exploration in each side $N_{E}$.
\vspace{2mm}
\State \textbf{Out:}
The final planable grasping directions of two arms $d_{L}$, $d_{R}$.
\vspace{2mm}
\State
Compute the ideal grasping directions  $D_{L}$, $D_{R}$.

\If {$D_{L}$, $D_{R}$ is planable}
\State
$d_{L}$ = $D_{L}$, $d_{R}$ = $D_{R}$;
\State
\Return $d_{L}$, $d_{R}$
\EndIf

\State
Set the minimal whole error of two arms $e_{min}$ = $\infty$

\For{$d_{l} = 0$; $ d_{l} \leqslant d_{I} \times N_{E}$; $d_{l} = d_{l}+d_{I}$}
\For{$d_{r} = 0$; $ d_{r} \leqslant d_{I} \times N_{E}$; $d_{r} = d_{r}+d_{I}$}
\State
Compute the error of left arm and right arm, $e_{l}$ = $d_{l}/d_{I}$, $e_{r}$ = $d_{r}/d_{I}$;
\State
Compute the whole error $e_{lr}$ = $e_{l} \times e_{r}$;

\If{$d_{l}$, $d_{r}$ is planable $and$ $e_{lr}<e_{min}$}
\State
$d_{L}$ = $d_{l}$; $d_{R}$ = $d_{r}$;
$e_{min}$ = $e_{lr}$;
\EndIf
\EndFor
\EndFor

\vspace{2mm}
\State \Return $d_{L}$, $d_{R}$
\end{algorithmic}
\caption{The Pose Exploration Algorithm for Planing Dual-Arms Grasping.}
\end{algorithm}

%% file: exp_overview.tex

%% file: grasping_exp.tex
\begin{table*}[t]
\small
\centering
\caption{\label{tab:grasping1}The grasping success rate on different types of clothing.}
\begin{tabular}{|p{1.6cm}|p{2.2cm}|p{2.2cm}|p{2.2cm}|p{2.2cm}|p{2.2cm}|p{1.2cm}|}
\hline
Successful Rate&~~~t-shirts&~~~shirts&~~~sweaters&~~~jeans&~~~jackets&average\\
\hline
categories & ~~~~90.0\% & ~~~~78.3\% & ~~~~93.3\% & ~~~~76.7\% & ~~~~85.0\% & 84.7\%\\
\hline
items &95\%$|$85\%$|$90\%&70\%$|$80\%$|$85\%&95\%$|$95\%$|$90\% &70\%$|$75\%$|$85\%&80\%$|$95\%$|$80\%& \\
\hline
\end{tabular}
\end{table*}

\begin{table*}[t]
\small
\centering
\caption{\label{tab:grasping2}The required number of grasping shots for a successful grasping on different types of clothing.}
\begin{tabular}{|p{1.8cm}|p{1.8cm}|p{1.8cm}|p{1.6cm}|p{1.8cm}|p{1.8cm}|p{1.8cm}|}
\hline
Number of Shots&~~~t-shirts&~~~shirts&~~~sweaters&~~~jeans&~~~jackets&~~~average\\
\hline
categories& ~~~~1.1 & ~~~~1.23 & ~~~~1.1 & ~~~~1.27 & ~~~~1.17 & ~~~~1.17\\
\hline
items& 1$|$1.2$|$1.1 & 1.2$|$1.2$|$1.3& 1.1$|$1.1$|$1.1 & 1.4$|$1.3$|$1.1 & 1.1$|$1.1$|$ 1.3 & ~~~~|\\
\hline
\end{tabular}
\end{table*}

In this experiments, the grasping performance of the proposed approach is evaluated. The evaluation of robotic grasping has two parts: firstly, the success rate of single-shot grasping is investigated; secondly, the effectiveness of grasping is evaluated by counting the required number of shots for completing a successful grasping.

\subsubsection{Single-Shot Grasping Experiment}
In the first grasping experiment, the grasping performance among five categories including t-shirts, shirts, sweaters, jeans and jackets are tested. Each category has three items of clothing, and 20 grasping experiments are tested on each item of clothing (in total 300 experiments). In each grasping experiment, the selected item of clothing is initialized to an arbitrary configuration by grasping and dropping it on the table. A successful grasping case means that: the gripper is moved to the position indicated by the visual feature; the grasping pose fits the shape of the region to grasp; and the clothing is fetched up. Since this work is focused on visual perception of grasping rather than kinematics, in these experiments, the flatness-priority grasping is carried out first. If the inverse-kinematics cannot be solved, the height-priority strategy is then applied (introduced in Section \ref{sec:grasping_strategy}).

The experimental results of the first grasping experiment are shown in Table \ref{tab:grasping1}. Overall, the grasping success rate varies from 76.7\% to 93.3\% on different types of clothing. This difference can be attributed into the difference of clothes materials. In other words, the thickness and stiffness variation of clothes' materials brings different challenges to grasping. Specifically, the sweaters and t-shirts performed the best (93.3\%,90\%) while jeans and shirts obtained the lowest scores (78.3\%,76.7\%). The reason is two-fold: firstly, the more stiff the clothing material is, the more difficult the grasping is; and also, the more wrinkles the clothing configuration has, the easier the grasping is. On average, the proposed method is able to achieve 84.7\% success rate among the five categories of clothing. In addition, the grasping performance on each item of clothing is also shown in the table. All 15 items of test clothing can achieve at least 70\% successful grasping rate.

\subsubsection{Multiple-Shot Grasping Experiment}
Apart from the single-shot grasping success rate, the other criterion required to be evaluated is the number of trails for each completed grasping. The latter allowed us to demonstrate that our visual architecture and extracted features is able to handle difficult configurations\footnote{\kevinupdate{Here, difficult configurations mean those lack of graspable positions. They often appear in clothes of stiff fabric e.g. shirt and jeans.}}. In our implementation, the proposed grasping feature provides a ranked array of grasping candidates, then the robot attempts to grasp them sequentiality until the grasping is completed successfully. In order to acquire the grasping status, tactile sensors are used to detect whether the gripper is holding the garment.

Experimental results are shown in Table \ref{tab:grasping2}, in which 150 successful graspings are completed (10 experiments on each item of clothing) and 1.17 trails are required for each successful grasping on average. As shown in the table, similarly to the first grasping experiments, stiff clothes such as jeans and shirts require more grasping trails (1.27 and 1.23 times, respectively). The robot requires the least number of grasping trails on sweaters and t-shirts (1.1 and 1.1 times, respectively). The deviation between different items of clothing is small; the required number of trails ranges from 1.0 to 1.4 among all of the items of clothing. Among these 150 successful graspings, only 1 grasping is completed after 4 attempts, 3 graspings after 3 attempts, 17 graspings after 2 attempts, and the remainubf 129 graspings are completed on the first attempt.

Overall, the experimental results of the proposed grasping method demonstrate a reliable grasping performance in terms of its grasping success rate (84.7\%) and its effectiveness of grasping difficult configurations (1.17 trails on average.

%% file: flattening_exp.tex
This section evaluates the performance of the proposed visual perception architecture on localising and quantifying wrinkles, and the integrated autonomous dual-arm flattening. This evaluation comprises three different experiments. Firstly, a benchmark flattening experiment comprising eight tasks is established to verify the performance and reliability for flattening a single wrinkle using dual-arm planning (Section \ref{sec:flattening_exp1}). While, in Section \ref{sec:flattening_exp2}, the second experiment demonstrates the performance of the proposed approach while flattening a highly wrinkled garment, comparing our robot stereo head system with standard Kinect-like cameras. Finally, Section \ref{sec:flattening_exp3} investigates the adaptability of the proposed flattening approach on different types of clothing, in which the performance of flattening towels, t-shirts and shorts are evaluated and compared.

The proposed visual perception architecture is able to detect wrinkles that are barely discernible to human eyes unless close inspection on the garment is carried out. As it is not necessary to flatten these wrinkles, a halting criterion is therefore proposed, which scores the amount of `flatness' based on the amount of the pulling distance computed in Eq. \ref{eq:flattening_dist}. In these experiments, if the flattening distances inferred by the detected wrinkles are less than 0.5 cm (barely perceptible), the garment is considered to be flattened\footnote{This value is obtained by averaging manually flattened garment examples performed by a human user.}.

\subsubsection{Benchmark Flattening}\label{sec:flattening_exp1}

\begin{figure}[thpb]
\centering
\includegraphics[width= 0.4\textwidth,natwidth=446,natheight=880]{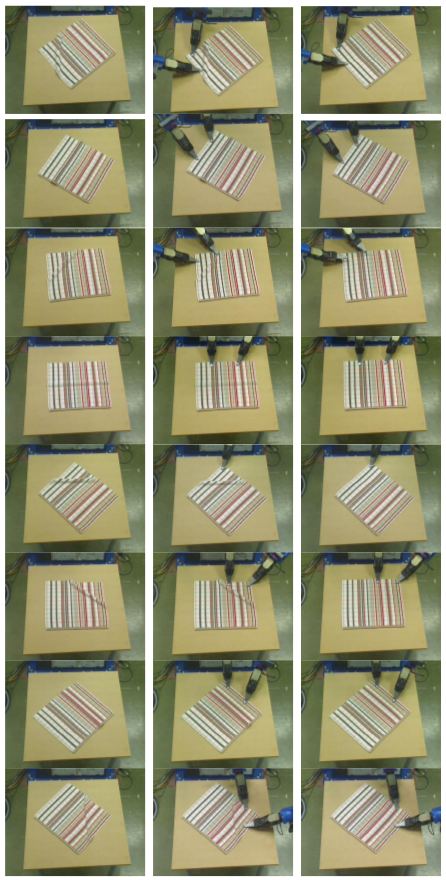}

\caption{Eight benchmark experiments on a single wrinkle using dual-arm planning. Each row depicts an experiment, in which the left images show the stage before flattening; middle, during flattening; and right, after flattening.}
\label{fig:bench-mark}
\end{figure}

\begin{table*}[thpb]
\small
\centering
\caption{\label{tab:bench-mark}The Required Number of Iterations (RNI) in the experiments.}

\begin{tabular}{|p{3.5cm}|p{0.7cm}|p{0.7cm}|p{0.7cm}|p{0.7cm}|p{0.7cm}|p{0.7cm}|p{0.75cm}|p{0.7cm}|p{1.2cm}|p{0.9cm}|}
\hline
Benchmark Experiments &exp1&exp2&exp3&exp4&exp5&exp6&exp7&exp8&average\\
\hline
RNI & 1 & 1 & 1 & 1 & 1 & 1 & 1 & 1 & 1 \\
\hline
Dual-Arm Success Rate & 100\% & 100\% & 80\% & 100\% & 0\% & 100\% & 100\% & 100\% & 85\% \\
\hline
Grasping Success Rate & 100\% & 100\% & 100\% & 100\% & 100\% & 100\% & 100\% & 100\% & 100\% \\
\hline
\end{tabular}
\end{table*}

\begin{table*}[thpb]
\small
\centering
\caption{\label{tab:comparison_arm_sensor}The Required Number of Iterations (RNI) for flattening in highly wrinkled experiments. See text for a detailed description.}

\begin{tabular}{|p{1.9cm}|p{0.5cm}|p{0.6cm}|p{0.6cm}|p{0.6cm}|p{0.6cm}|p{0.6cm}|p{0.6cm}|p{0.6cm}|p{0.6cm}|p{0.75cm}|p{1.0cm}|p{0.6cm}|p{1.4cm}|}
\hline
Flattening Experiments &exp1&exp2&exp3&exp4&exp5&exp6&exp7&exp8&exp9&exp10&AVE&STD&Dual-Arm Success\\
\hline
Dual-Arm (RH) & 4(4) & 5(4) & 6(4) & 5(4) & 4(3) & 5(3) & 4(2) & 5(2) & 3(2) & 6(3) & \bf{4.7(3.1)} & \bf{0.95} & \bf{65.9\%}\\
\hline
Dual-Arm (Xtion) & 7(4) & 8(4) & 7(3) & 12(4) & 8(4) & 13(7) & 11(3) & 10(5) & 9(5) & 10(5) & 9.5(4.4) & 2.07 & 46.3\%\\
\hline
Single-Arm (RH) & 7 & 12 & 5 & 8 & 7 & 7 & 12 & 14 & 8 & 6 & 8.6 & 2.99 & -\\
\hline
Single-Arm (Xtion) & 10 & 12 & 17 & 11 & 12 & 19 & 13 & 12 & 11 & 14 & 13.1 & 2.85 & -\\
\hline
\end{tabular}
\end{table*}

\begin{table*}[t]
\small
\centering
\caption{\label{tab:comparison_type}The Required Numbers of Iterations (RNI) for flattening different types of garments.}

\begin{tabular}{|p{3.5cm}|p{0.6cm}|p{0.6cm}|p{0.6cm}|p{0.6cm}|p{0.6cm}|p{0.6cm}|p{0.6cm}|p{0.6cm}|p{0.6cm}|p{0.6cm}|p{0.7cm}|p{0.7cm}|}
\hline
RNI of tasks &exp1&exp2&exp3&exp4&exp5&exp6&exp7&exp8&exp9&exp10&AVE&STD\\
\hline
flattening towels & 4 & 5 & 6 & 5 & 4 & 5 & 4 & 5 & 3 & 6 & 4.7 & 0.95 \\
\hline
flattening t-shirt & 5 & 7 & 12 & 11 & 7 & 8 & 12 & 9 & 12 & 6 & 8.9 & 2.68 \\
\hline
flattening pants(shorts) & 11 & 10 & 5 & 14 & 4 & 3 & 4 & 3 & 2 & 7 & 6.3 & 4.05 \\
\hline
\end{tabular}
\end{table*}

\begin{figure*}[thpb]
  \centering
	\includegraphics[width= 1\textwidth,natwidth=967,natheight=617]{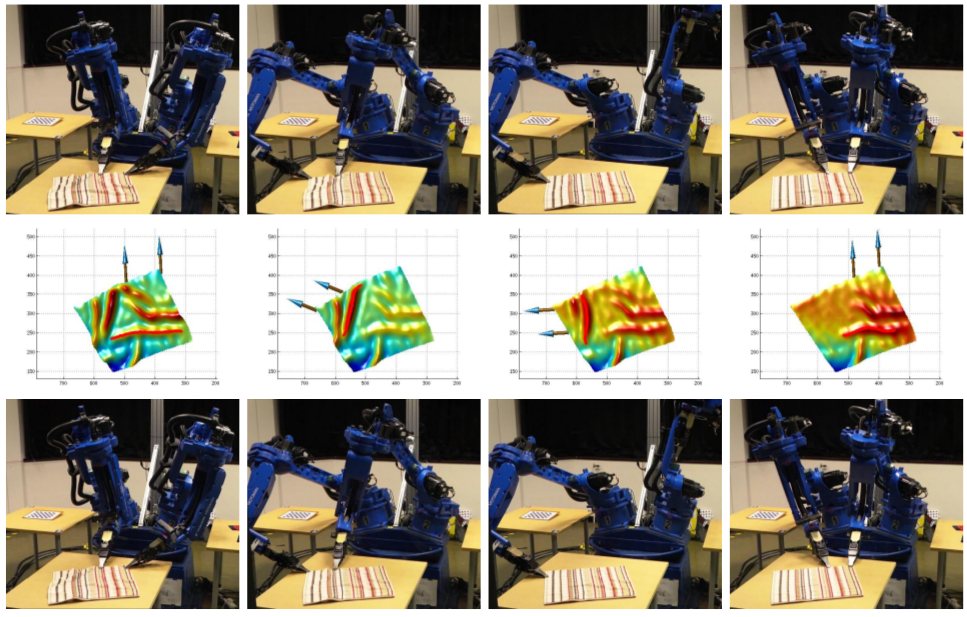}
    \caption{\label{fig:demo1} A demonstration of flattening an item of highly wrinkled towel. Each column depicts one iteration in the experiment. The top row depicts the towel state before the iteration; middle row, the detected largest wrinkles and the inferred forces; bottom row, the towel state after the iteration. On the third iteration, dual-arm planing demonstrated infeasible to execute, so a single-arm manoeuvre is formulated and applied.}
\end{figure*}

\begin{figure*}[thpb]
  \centering
	\includegraphics[width= 1\textwidth,natwidth=969,natheight=345]{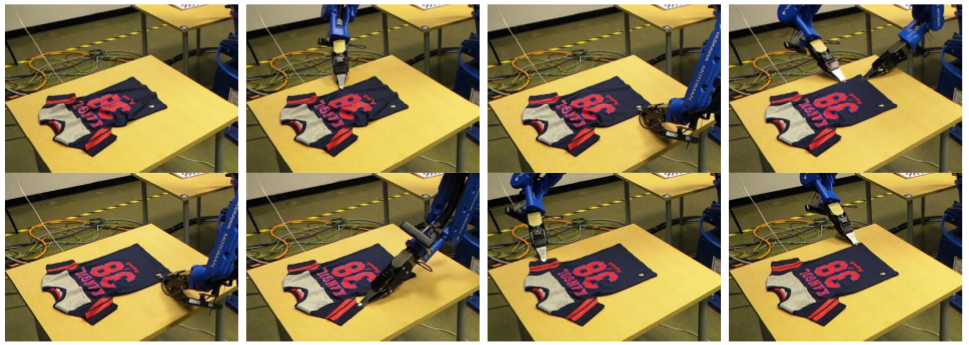}
    \caption{\label{fig:demo2} An example of flattening a T-shirt. As it is observed, the proposed flattening approach is able to adapt to any shape of garment, the robot can grasp the sleeves and stretch the wrinkles successfully. }
\end{figure*}

The aim of the first experiment is to evaluate the performance of the proposed flattening method under pre-defined single wrinkle configurations as well as the dual-arm planning performance for flattening in different directions. For this purpose, eight benchmark flattening experiments are performed. As shown in Fig. \ref{fig:bench-mark}, in each instance there is one salient wrinkle distributed in the range of 45 degrees to -45 degrees (from the robot's view). In order to obtain a stable evaluation, each experiment is repeated 5 times, and results are shown in Table \ref{tab:bench-mark}.

It can be deduced from Table \ref{tab:bench-mark} that the proposed flattening approach is able to flatten these eight benchmark experiments with only one iteration. Moreover, the success rate for dual-arm planning is 85\%, where the robot successfully grasps the edge(s) of the garment in all of these experiments. Experiment 5 shows a failed case while using both arms; this is because the robot reaches the limitation of its joints and the inverse kinematic planner adopted.

\subsubsection{Highly-Wrinkled Towel Flattening}\label{sec:flattening_exp2}

In order to investigate the contribution of the depth sensing provided by our stereo head and dual-arm manipulation strategy within the proposed approach in terms of autonomous flattening of highly wrinkled garments, the flattening performance between single-arm and dual-arm strategies is compared. Similarly, in order to demonstrate the utility of high-quality sensing capabilities for the dexterous clothes manipulation, the flattening performance between our stereo robot head and a Kinect-like sensor (here the ASUA Xtion PRO is used\footnote{\url{http://www.asus.com/uk/Multimedia/Xtion_PRO_LIVE/}}) is compared.

Therefore, for each experiment, a square towel is randomly wrinkled - wrinkles are distributed in different directions without following an order. Then different flattening strategies are applied (single-arm or dual-arm) with either the robot stereo head or Xtion. For comparison, 4 groups of experiments are carried out: (1) dual-arm using robot head, (2) single-arm using robot head, (3) dual-arm using Xtion and (4) single-arm using Xtion. To measure the overall performance and reliability, 10 experiments are conducted for each group of experiment and the required number of iterations (RNI) is counted as shown in Table \ref{tab:comparison_arm_sensor}. In Table \ref{tab:comparison_arm_sensor}, each column represents the experiment index for each of the groups proposed above. Values in parentheses show the RNI where dual-arm planning was successful while the rest of the values show the RNI for each experiment.

As shown in Table \ref{tab:comparison_arm_sensor}, the average RNI for dual-arm flattening using robot head is 4.7 (achieving 65.9\% arm planning success rate) while single-arm is 8.6. The average RNI for dual-arm flattening using Xtion is 9.5 (achieving 46.3\% dual-arm planning success rate), while single-arm is 13.1. This result shows that a dual-arm strategy achieves a much more efficient performance on flattening than a signal-arm strategy. The standard deviation (STD) of each group of experiments is also calculated, where the STD for dual-arm flattening is 0.95 (using robot head) and 2.07 (using Xtion) while for single-arm is 2.99 (using robot head) and 2.85 (using Xtion). As expected, a dual-arm strategy is not only more efficient but also more stable than a single-arm strategy. Likewise, from the sensors' perspective, the robot is able to complete a flattening task successfully within 4.7 iterations (dual-arm case) using the stereo robot head as opposed to 9.5 iterations while using Xtion. Overall, the our robot head clearly outperforms the Xtion in both dual-arm flattening and single-arm flattening experiments.

The results described above demonstrate that the dual-arm strategy is more efficient in flattening long wrinkles than the single-arm because the latter approach usually breaks long wrinkles into two short wrinkles. Likewise, comparing the our stereo head and Xtion, as observed during the experiments, it is difficult to quantify the wrinkles and also estimate the accurate flattening displacement (especially for small wrinkles) from the Xtion depth data because the depth map is noisier than the robot head (the high frequency noise is usually more than 0.5cm). Furthermore, long wrinkles captured by Xtion are often split into two small wrinkles due to the poor quality of the depth map, which in turn results in more flattening iterations (the short detected wrinkles are likely to have a lower dual-arm planing success rate). 

\subsubsection{Flattening Different Types of Garments}\label{sec:flattening_exp3}

\begin{figure*}[thpb]
\centering
	\includegraphics[width= 1\textwidth,natwidth=1129,natheight=184]{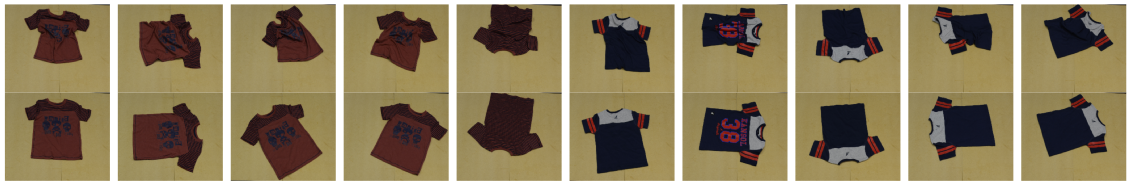}
\caption{Ten experiments of flattening t-shorts. Each column demonstrates a flattening experiment, in which the upper image refers to the initial configuration and the lower final configuration.}
\label{fig:flattening_tshirt}
\end{figure*}

\begin{figure*}[thpb]
\centering
	\includegraphics[width= 1\textwidth,natwidth=1129,natheight=184]{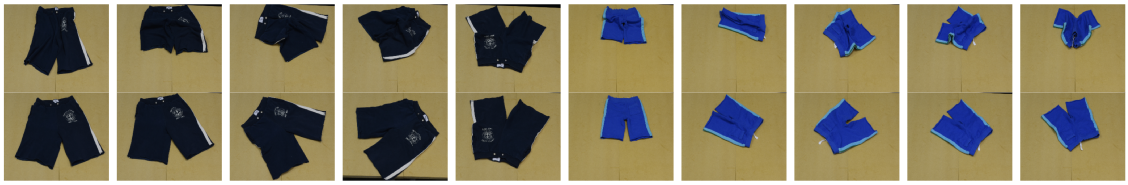}
\caption{Ten experiments of flattening shorts. Each column demonstrates a flattening experiment, in which the upper image refers to the initial configuration and the lower final configuration.}
\label{fig:flattening_pant}
\end{figure*}

Since the proposed flattening approach has no constraints on the shape of the garment, this section evaluates the performance of this method on flattening other types of clothing, namely t-shirts and shorts. Ten flattening experiments are performed for each type of clothing. Examples are shown in Fig. \ref{fig:flattening_tshirt} and Fig. \ref{fig:flattening_pant}, respectively.

The RNIs of different clothes categories are shown in Table \ref{tab:comparison_type}, and here the towel flattening performance is presented as the baseline performance. As shown in the table, towels require an average of only 4.7 iterations to complete flattening. Shorts need more iterations on average(6.3) and t-shirts require still more (8.9). The reason is that towels are of the simplest shape among these tree categories of clothing, while the shape of shorts is more complicated and that of t-shirts is the most complex. This experimental result demonstrates that the proposed approach is able to flatten different categories of clothing and that the RNI of flattening clothing is propagating to the complexities of the clothing's 2D topological shape.

\subsection{Summary}
The proposed autonomous grasping is evaluated in both single-trial and interactive-trial experiments, showing robustness among the clothes types. And the validation of the reported autonomous flattening behaviours has been undertaken and has demonstrated that dual-arm flattening requires significantly fewer manipulation iterations than single-arm flattening. The experimental results also indicate that the dexterous clothes operation (such as flattening) is significantly influenced by the quality of the RGB-D sensor $-$ using a customized off-the-shelf high-resolution stereo-head outperforms the commercial low-resolution Kinect-like cameras in terms of required number of flattening iterations (RNIs).

%% file: conclusions.tex
In this paper, a novel visual perception architecture is proposed for clothes configuration parsing, and this architecture is integrated with an active stereo vision system and dual-arm CloPeMa robot to demonstrate dexterous garment grasping and flattening. The proposed approach is based on generic 3D surface analysis, and tend to fully understand the landmark structures distributed on the clothing surface, thereby demonstrating the adaptability for multiple visually-guided clothes manipulation tasks. From the experimental validation, the conclusions are: firstly, the proposed visual perception architecture is able to parse the various garment configurations by detecting and quantifying structures i.e. grasping triplets and wrinkles; secondly, the stereo robot head used in this research outperforms Kinect-like depth sensors in terms of dexterous visually-guided garment manipulation; finally, the proposed dual-arm flattening strategy greatly improves garment manipulation efficiency as compared to the single-arm strategy. The integrated stereo head, visual perception architecture and visually-guided manipulation systems demonstrate the effectiveness of grasping and flattening different types of garments.  On the other hand, the integrated autonomous flattening employs the perception-manipulation cycles, and consequently the clothing configuration is modified towards the flattening goal.